\documentclass[11pt,a4paper]{article}
\usepackage[hyperref]{emnlp2020}
\usepackage{times}
\usepackage{latexsym}
\usepackage{amsmath}
\usepackage{flushend}

\usepackage{microtype}
\usepackage{tikz}
\def\checkmark{\tikz\fill[scale=0.4](0,.35) -- (.25,0) -- (1,.7) -- (.25,.15) -- cycle;} 

\aclfinalcopy 

\usepackage{lineno}
\usepackage{booktabs}
\usepackage{multirow}
\usepackage{natbib}
\usepackage{enumitem}
\usepackage{balance}
\usepackage{flushend}

    \makeatletter
\def\@fnsymbol#1{\ensuremath{\ifcase#1\or \dagger\or \ddagger\or
   \mathsection\or \mathparagraph\or \|\or **\or \dagger\dagger
   \or \ddagger\ddagger \else\@ctrerr\fi}}
    \makeatother

\title{A Wrong Answer or a Wrong Question?\\An Intricate Relationship between Question Reformulation and Answer Selection in Conversational Question Answering}

\author{Svitlana Vakulenko\thanks{\hspace{2mm}Work done as an intern at 
Apple Inc.} \\
  University of Amsterdam  \\
  \texttt{s.vakulenko@uva.nl} \\\And
  Shayne Longpre, Zhucheng Tu, Raviteja Anantha \\
  Apple Inc.\\
  \texttt{\{slongpre, zhucheng\_tu,} \\
  \texttt{raviteja\_anantha\}@apple.com}}

\begin{document}
\maketitle
\begin{abstract}
The dependency between an adequate question formulation and correct answer selection is a very intriguing but still underexplored area.
In this paper, we show that question rewriting (QR) of the conversational context allows to shed more light on this phenomenon and also use it to evaluate robustness of different answer selection approaches.
We introduce a simple framework that enables an automated analysis of the conversational question answering (QA) performance using question rewrites, and present the results of this analysis on the TREC CAsT and QuAC (CANARD) datasets.
Our experiments uncover sensitivity to question formulation of the popular state-of-the-art models for reading comprehension and passage ranking.
Our results demonstrate that the reading comprehension model is insensitive to question formulation, while the passage ranking changes dramatically with a little variation in the input question.
The benefit of QR is that it allows us to pinpoint and group such cases automatically.
We show how to use this methodology to verify whether QA models are really learning the task or just finding shortcuts in the dataset, and better understand the frequent types of error they make.
\end{abstract}
\maketitle

\section{Introduction}

Conversational question answering (QA) is a new and important task which allows systems to advance from answering stand-alone questions to answering a sequence of related questions~\cite{DBLP:conf/emnlp/ChoiHIYYCLZ18,DBLP:journals/tacl/ReddyCM19,dalton2019trec}. Such sequences contain questions that usually revolve around the same topic and its subtopics, which is also common for a human conversation. The most pronounced characteristics of such question sequences are anaphoric expressions and ellipsis, which make the follow-up questions ambiguous outside of the conversation context. For example, consider a question ``\textit{When was \textbf{it} discovered?}''. It is not possible to answer the question without resolving the pronoun \textbf{it} (example of an anaphoric expression). Ellipsis are even harder to resolve since they omit information without leaving any references. For example, a question ``\textit{When?}'' can naturally follow an answer to the previous question, such as ``\textit{Friedrich Miescher discovered DNA.}'' 

Question rewriting (QR) was recently introduced as an independent component for conversational QA~\cite{elgohary2019can,ours,DBLP:conf/sigir/YuLYXBG020}.
Query rewriting received considerable attention in the information retrieval community before but not in a conversational context~\cite{he2016learning}.
\citet{ren2018conversational} performed similar experiments using query reformulations mined from search sessions.
However, half of their samples were keyword queries rather than natural language questions and their dataset was not released to the community.
In this paper, we show that QR is not only operational in extending standard QA models to the conversational scenario but can be also used for their evaluation.

An input to the QR component is a question and previous conversation turns. The QR component is designed to transform all ambiguous questions, e.g., ``\textit{When?}'', into their unambiguous equivalents, e.g., ``\textit{When was DNA discovered?}''.
Such unambiguous questions can be then processed by any standard QA model outside of the conversation history. 


Clearly, the quality of question formulation interacts with the ability to answer this question. In this paper, we show that introducing QR component in the conversational QA architecture, by decoupling question interpretation in context from the question answering task, provides us with a unique opportunity to gain an insight on the interaction between the two tasks. Ultimately, we are interested in whether this interaction can potentially help us to improve the performance on the end-to-end conversational QA task. To this end we formulate our main research question:

\textit{How do differences in question formulation in a conversational setting affect question answering performance?}

The standard approach to QA evaluation is to measure model performance on a benchmark dataset with respect to the ground-truth answers, such as text span overlap in the reading comprehension task, or NDCG@3 in the passage retrieval task.
However, such evaluation setups may also have their limitations with respect to the biases in the task formulation and insufficient data diversity that allow models to learn shortcuts and overfit the benchmark dataset~\cite{DBLP:journals/corr/abs-2004-07780}.
There is already an ample evidence of the pitfalls in the evaluation setup of the reading comprehension task highlighting that the state-of-the-art models tend to learn answering questions using superficial clues~\cite{DBLP:conf/emnlp/JiaL17,DBLP:conf/acl/SinghGR18,DBLP:conf/iclr/LewisF19}.

We aim to further contribute to the research area studying robustness of QA models by extending the evaluation setup to the conversational QA task. To this end, we introduce an error analysis framework based on conversational question rewrites. The goal of the framework is to evaluate robustness of QA models using the inherent properties of the conversational setup itself.
More specifically, we contrast the results obtained for ambiguous and rewritten questions outside of the conversation context. We show that this data is very well suited for analysing performance and debugging QA models. In our experiments, we evaluate two popular QA architectures proposed in the context of the conversational reading comprehension (QuAC)~\cite{DBLP:conf/emnlp/ChoiHIYYCLZ18} and conversational passage ranking (TREC CAsT)~\cite{dalton2019trec} tasks.

Our results show that the state-of-the-art models for passage retrieval are rather sensitive to differences in question formulation. On the other hand, the models trained on the reading comprehension task tend to find correct answers even to incomplete ambiguous questions. We believe that these findings can stimulate more research in this area and help to inform future evaluation setups for the conversational QA tasks.


\section{Experimental Setup}
\label{sec:setup}

To answer our research question and illustrate the application of the proposed evaluation framework in practice, we use the same experimental setup introduced in our earlier work~\cite{ours}. This architecture consists of two independent components: QR and QA. It was previously evaluated against competitive approaches and baselines setting the new state-of-the-art results on the TREC CAsT 2019 dataset. The QR model was also shown to improve QA performance on both passage retrieval and reading comprehension tasks.

In the following  subsection, we describe the datasets, models, and metrics that were used in our evaluation. Note, however, that our error analysis framework can be applied to other models and metrics as well. The only requirement for applying the framework is the QR-QA architecture that provides two separate outputs in terms of question rewrites and answers. We show that the same framework can be applied for both reading comprehension and passage retrieval tasks, although they produce different types of answers and require different evaluation metrics.

\subsection{Datasets}

We chose two conversational QA datasets for the evaluation of our approach: (1) TREC CAsT for conversational passage ranking~\cite{dalton2019trec}, and (2)  CANARD, derived from Question Answering in Context (QuAC) for conversational reading comprehension~\cite{DBLP:conf/emnlp/ChoiHIYYCLZ18}. Since TREC CAsT is a relatively small dataset, we used only CANARD for training our QR model. The same QR model trained on CANARD was then evaluated on both CANARD and TREC CAsT independently. 

Following the setup of the TREC CAsT 2019, we use the MS MARCO passage retrieval \cite{DBLP:conf/nips/NguyenRSGTMD16} and the TREC CAR \cite{dietz2018trec} paragraph collections. After de-duplication, the MS MARCO collection contains 8.6M documents and the TREC CAR -- 29.8M documents. The model for passage retrieval is tuned on a sample from the MS MARCO passage retrieval dataset, which includes relevance judgements for 12.8M query-passage pairs with 399k unique queries~\cite{nogueira2019passage}. We evaluated on the test set with relevance judgements for 173 questions across 20 dialogues (topics).

We use CANARD~\cite{elgohary2019can} and QuAC~\cite{DBLP:conf/emnlp/ChoiHIYYCLZ18}  datasets jointly to analyse performance on the reading comprehension task. CANARD is built upon the QuAC dataset by employing human annotators to rewrite original questions from QuAC dialogues into explicit questions. CANARD contains 40.5k pairs of question rewrites that can be matched to the original answers in QuAC. We use CANARD splits for training and evaluation. Each answer in QuAC is annotated with a Wikipedia passage from which it was extracted alongside the correct answer spans within this passage. We use the question rewrites provided in CANARD and passages with answer spans from QuAC. In our experiments, we refer to this joint dataset as CANARD for brevity. Our model for reading comprehension was also pre-trained using MultiQA dataset~\cite{fisch2019mrqa} to further boost its performance. MultiQA contains 75k QA pairs from six standard QA benchmarks.

\subsection{Conversational QA Architecture}

Our architecture for conversational QA is designed to be modular by separating the original task into two subtasks. The subtasks are  (1) QR, responsible for the conversational context understanding and question formulation, and (2) QA that exactly corresponds to the standard non-conversational task of answer selection (reading comprehension or passage retrieval). Therefore, the output of the QR component is an unambiguous question that is subsequently used as input to the QA component to produce the answer.

\paragraph{Question Rewriting}

The task of question rewriting is to reformulate every follow-up question in a conversation, such that the question can be unambiguously interpreted without accessing the conversation history. The input to the QR model is the question with previous conversation turns separated with a $[SEP]$ token (in our experiments, we use up to maximum of 5 previous conversation turns). Using our running example, the input would be: \textit{Friedrich Miescher discovered DNA $[SEP]$ When?} and the expected output from the QR model is: \textit{When was DNA discovered?}.

Our QR model is based on a unidirectional Transformer (decoder) designed for the sequence generation task. It was initialised with the weights of the pre-trained GPT2~\cite{radford2019language} and further fine-tuned on the QR task. The training objective in QR is to predict the output sequence as in the ground truth question rewrites produced by human annotators. The model is trained via the teacher forcing approach. The loss is calculated with the negative log-likelihood (cross-entropy) function. At inference time, the question rewrites are generated recursively turn by turn for each of the dialogues using the previously generated rewrites as input corresponding to the dialogue history, i.e., previous turns.

For our experiments, we adopt the same QR model architecture proposed in the previous work (Transformer++)~\cite{qrecc,ours}.
It was shown to outperform a co-reference baseline and other Transformer-based models on CANARD, TREC CAsT and QReCC.


\paragraph{Question Answering}

We experiment with two different QA models that reflect the state-of-the-art in reading comprehension and passage retrieval~\cite{nogueira2019passage}. All our QA models are initialised with a pre-trained $BERT_{LARGE}$~\cite{devlin2018bert} and then fine-tuned on each of the target tasks.

Our model for reading comprehension follows the standard architecture design for this task. We restrict our implementation to the simplest but a very competitive model architecture that more complex approaches usually build upon~\cite{liu2019roberta,lan2019albert}. This model consists of a Transformer-based bidirectional encoder and an output layer that predicts the answer span. The input to the model is the sequence of tokens formed by concatenating a question and a passage, the two are separated with a $[SEP]$ token. 

Our passage retrieval approach is implemented following \citet{nogueira2019passage}. It uses Anserini for the candidate selection phase with BM25 (top-1000 passages) and $BERT_{LARGE}$ for the passage re-ranking phase.\footnote{\url{https://github.com/nyu-dl/dl4marco-bert}} The re-ranking model was tuned on a sample from MS MARCO with 12.8M query-passage pairs and 399k unique queries.

\subsection{Metrics}

We use the standard performance metrics for each of the QA subtasks.
We use normalized discounted cumulative gain (\textit{NDCG@3}) and precision on the top-passage (\textit{P@1}) to evaluate quality of passage retrieval (with a relevance threshold of 2 in accordance with the official evaluation guidelines for CAsT). We use \textit{F1} metric for reading comprehension, which measures word overlap between the predicted answer span and the ground truth.

In contrast with QA, there is no established methodology for reporting the QR performance yet.
Existing studies tend to report performance using BLEU metrics following the original CANARD paper~\cite{DBLP:conf/sigir/YuLYXBG020,lin2020query}.

We conducted a systematic evaluation of different performance metrics to find a subset that correlates with the human judgement of the quality of question rewrites (see more details in \cite{qrecc}). 
Our analysis showed that ROUGE-1 Recall (ROUGE-1 R)~\cite{lin2004rouge} and Universal Sentence Encoder (\textit{USE})~\cite{DBLP:conf/emnlp/CerYKHLJCGYTSK18} embeddings correlate with the human judgement of the rewriting quality the most (Pearson 0.69 for ROUGE-1 R and Pearson 0.71 for USE).
Therefore, we also use ROUGE-1 R and USE in this study to measure similarity between the model rewrites and the ground truth.

ROUGE is traditionally employed for evaluation of the text summarisation approaches.
While ROUGE is limited to measuring lexical overlap between the two input texts, the USE model outputs dense vector representations that are designed to indicate semantic similarities between sentences beyond word overlap.

\begin{table}[t]
    	\caption{Break-down analysis of the passage retrieval on TREC CAsT. Each row represents a group of QA samples that exhibit similar behaviour. We consider three types of input for every QA sample: the question from the test set (Original), generated by the best QR model (Transformer++) or rewritten manually (Human). The numbers correspond to the count of QA samples for each of the groups. The numbers in parenthesis indicate how many questions in the ground truth do not need rewriting, i.e., Human = Original.}
	\centering
	\label{table:trec_breakdown}
	\resizebox{\columnwidth}{!}{
		\begin{small}
			\begin{tabular}{ccccccc}
				\toprule
				 & & & P@1  & \multicolumn{3}{c}{NDCG@3} \\
				Original & QR & Human & = 1 & $>$ 0 & $\geq$ 0.5 & = 1 \\
				\midrule
				$\times$ & $\times$ & $\times$ & 49 (14) & 10 (1) & 55 (20) & 154 (49) \\ 
				$\checkmark$ & $\times$ & $\times$ & 0 & 0 & 0 & 0 \\
				$\times$ & $\checkmark$ & $\times$ & 2 & 0 & 1 & 0 \\
				$\checkmark$ & $\checkmark$ & $\times$ & 0 & 1 & 1 & 0 \\
				\midrule
				$\times$ & $\times$ & $\checkmark$ & 19 & 10 & 25 & 4 \\
				$\checkmark$ & $\times$ & $\checkmark$ & 0 & 1 & 0 & 0 \\
				$\times$ & $\checkmark$ & $\checkmark$ & 48 & 63 & 47 & 11 \\
				$\checkmark$ & $\checkmark$ & $\checkmark$ & 55 (37) & 88 (52) & 44 (33) & 4 (4)  \\ 
				\midrule
				 \multicolumn{3}{c}{} & Total & \multicolumn{3}{r}{173 (53)} \\
				\bottomrule
			\end{tabular}
	\end{small}
	}
\end{table}

\begin{table}[t]
\caption{Break-down analysis of all reading comprehension results for the CANARD dataset, similar to Table~\ref{table:trec_breakdown}. Observe that Table~\ref{table:trec_breakdown} is much more sparse than Table~\ref{table:quac_breakdown}. There are almost no cases for which original or model-rewritten questions outperformed human rewriting when ranking passages. On the contrary, Table~\ref{table:quac_breakdown} indicates a considerable number of anomalous cases in which the reading comprehension model was able to answer incomplete follow-up questions but failed in answering ground-truth questions (rows 2-4).}
	\centering
	\label{table:quac_breakdown}
	\resizebox{\columnwidth}{!}{
		\begin{small}
			\begin{tabular}{cccccc}
				\toprule
				Original & QR & Human & F1 $>$ 0 & F1 $\ge$ 0.5 & F1 = 1  \\
				\midrule
				$\times$ & $\times$ & $\times$ & 847 (136) & 1855 (235) & 2701 (332) \\
				$\checkmark$ & $\times$ & $\times$ & 174 & 193 & 181 \\
				$\times$ & $\checkmark$ & $\times$ & 19 & 35 (2) & 40 (1) \\
				$\checkmark$ & $\checkmark$ & $\times$ & 135 & 153 & 120 \\
				\midrule
				$\times$ & $\times$ & $\checkmark$ & 141 & 288 & 232 \\
				$\checkmark$ & $\times$ & $\checkmark$ & 65 (1) & 57 (1) & 40 \\
				$\times$ & $\checkmark$ & $\checkmark$ & 226 & 324 & 269 \\
				$\checkmark$ & $\checkmark$ & $\checkmark$ & 3964 (529) & 2666 (428) & 1988 (333) \\
				\midrule
				 \multicolumn{3}{c}{} & Total & \multicolumn{2}{r}{5571 (666)} \\
				\bottomrule
			\end{tabular}
	\end{small}
	}
\end{table}

\begin{figure*}[h]
  \centering
    \includegraphics[width=\textwidth]{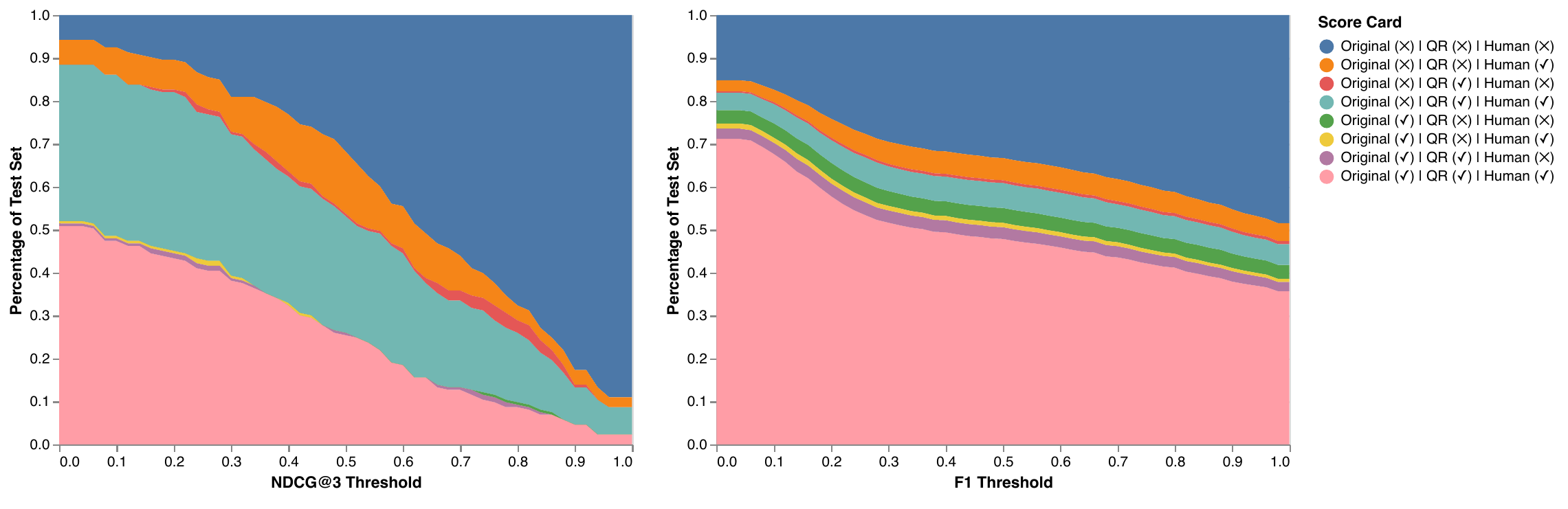}
    \caption{Break-down analysis for passage retrieval (left) and reading comprehension (right) results (best in color). This plot visualises the difference between the error distributions reported in Tables~\ref{table:trec_breakdown}-\ref{table:quac_breakdown} with a sliding cut-off threshold. The blue region at the top of the plots represents the proportion of errors in QA. The orange region represents the proportion of errors in QR. The light green region shows the proportion of samples that were both rewritten and answered correctly. The pink region at the bottom shows the importance of QR for the task. The larger the region the more questions were answered correctly without any rewriting. We observe that the difference between the two plots is very pronounced. The impact of QR is noticeable in passage retrieval, which is more sensitive to question formulation than the reading comprehension model.}
    \label{fig:trend_plots}
\end{figure*}

\section{Our Error Analysis Framework}
\label{sec:approach}

To trace how the difference in question formulation affects QA performance we compare the QR performance metrics with the QA performance metrics on the case-by-case basis, i.e., for each question-answer pair. Notice that we can apply the same approach for both retrieval and reading comprehension evaluation (see Table~\ref{table:quac_breakdown} for the results on reading comprehension).

Table~\ref{table:trec_breakdown} illustrates our approach. Each row of the table represents one of the combinations of the possible QA results for 3 types of question formulation. For every answer in a dataset we have 3 types of question formulation: (1) an original, possibly implicit, question (\textbf{Original}), (2) rewrites produced by the QR model (\textbf{QR}) and (3) rewrites produced by a human annotator (\textbf{Human}).

$\checkmark$ indicates that the answer produced by the QA model was correct or $\times$ -- incorrect, according to the thresholds provided in the right columns. For example, the first row of the table indicates the situation, when neither the original question, nor they generated or human rewrite were able to solicit the correct answer ($\times$ $\times$ $\times$). We then can automatically calculate how many samples in our results fall into each of these bins.

We assume human question rewrites as the ground truth. Therefore, all cases in which these rewrites did not result in a correct answer are errors of the QA component (rows 1-4 in Tables~\ref{table:trec_breakdown}-\ref{table:quac_breakdown}). The next two rows 5-6 show the cases, where human rewrites succeeded but the model rewrites failed, which we consider to be a likely error of the QR component. The last two rows are true positives for our model, where the last row combines cases where the original question was just copied without rewriting (numbers in brackets) and other cases when rewriting was not required.

Since there is no single binary measure for the answer correctness, we can pick different cut-off thresholds for the QA metrics. For example, P@1=1 will consider the answer correct if it came up at the top of the ranking; or F1=1 will consider the answer correct only in cases with full span overlap, i.e, exact matches only. Figure~\ref{fig:trend_plots} extends this analysis by considering all thresholds in the range [0; 1] with 0.02 intervals for NDCG@3 in retrieval and F1 in reading comprehension. This figure shows the proportion of different error types as well as the results sensitivity to the choice of the performance threshold.

\section{Evaluation Results}
\label{sec:results}

Our approach allows us to estimate that the majority of errors stem from the QA model: 29\% of the test samples for retrieval and 55\% for reading comprehension. 11\% of errors can be directly attributed to QR in the retrieval setup and 5\% in reading comprehension.

To estimate the impact of QR on QA, we consider only the last four rows in Table~\ref{table:quac_breakdown} for which QA model return a correct answer for Human questions.
Then, we divide the number of questions for which the Original question leads to the correct answer, i.e., without rewriting, by the total number of questions that can be answered correctly by our QA model (discarding the number of questions that were not rewritten by the annotators indicated in parenthesis).
For example, to estimate the number of questions correctly answered in CANARD (F1 = 1) without rewriting, i.e., using the Original question as input to the QA model:

\begin{align}
 \frac{40 + 1988 - 333}{232 + 40 + 269 + 1988 - 333} = 0.77
\end{align}

The proportion of QR errors for retrieval setup is higher than for reading comprehension setup.
In particular, we found that the majority of questions in CANARD test set (77\% F1 = 1) can be correctly answered using only the Original questions without any question rewriting, i.e., even when the questions are ambiguous.
For TREC CAsT, the chances of reaching the correct answer set using an ambiguous question are much lower (21\% P@1 = 1).
See Tables~\ref{table:trec_analysis}-\ref{table:quac_analysis} for the complete result set with different cut-off thresholds.



There are two anecdotal cases where our QR component was able to generate rewrites that helped to produce better ranking than the human-written questions. The first example shows that the re-ranking model does not handle paraphrases well. Original question: ``\textit{What are good sources in food?}'', human rewrite: ``\textit{What are good sources of \textbf{melatonin} in food?}'', model rewrite: ``\textit{What are good sources in food for \textbf{melatonin}}''. In the second example the human annotator and our model chose different context to disambiguate the original question. Original question: ``\textit{What about environmental factors?}'', human rewrite: ``\textit{What about environmental factors during the \textbf{Bronze Age collapse}?}'', model rewrite: ``\textit{What about environmental factors that lead to led to a \textbf{breakdown of trade}}''. Even though both model rewrites are not grammatically correct they solicited correct top-answers, while the human rewrites failed, which indicate flaws in the QA model performance.

\begin{table}[t]
\caption{The fraction of questions in TREC CAsT that were answered correctly without rewriting.}
	\centering
	\label{table:trec_analysis}
			\begin{tabular}{l cccc}
				\toprule
				 & P@1  & \multicolumn{3}{c}{NDCG@3} \\
				\multicolumn{1}{c}{Questions} & = 1 & $>$ 0 & $\geq$ 0.5 & = 1 \\
				\midrule
				All &  0.45 & 0.55 & 0.38 & 0.21 \\
				Human != Original & 0.21 & 0.34 & 0.13 & 0 \\
				\bottomrule
            \end{tabular}
\end{table}
\begin{table}[t]
\caption{The fraction of questions in CANARD that were answered correctly without rewriting.}
	\centering
	\label{table:quac_analysis}
			\begin{tabular}{l ccc}
				\toprule
				\multicolumn{1}{c}{Questions} & F1 $>$ 0 & F1 $\ge$ 0.5 & F1 = 1  \\
				\midrule
				All &  0.92 & 0.82 & 0.80 \\
				Human != Original &  0.91 & 0.79 & 0.77 \\
				\bottomrule
            \end{tabular}
\end{table}

\section{QA-QR Correlation}
\label{sec:correlation}

In this section, we check whether the QR metrics can predict the QA performance for the individual questions by measuring the correlation between the QR and QA metrics. This analysis shows how the change in question formulation affects the answer selection.
In other words, we are interested whether similar questions also produce similar answers, and whether distinct questions result in distinct answers.

To discover the correlation between question and answer similarity, we discarded all samples, where the human rewrites do not lead to the correct answers (top 4 rows in Tables~\ref{table:trec_breakdown}-\ref{table:quac_breakdown}). The remaining subset contains only the samples in which the QA model was able to find the correct answer. We then compute ROUGE for the pairs of human and generated rewrites, and measure its correlation with P@1 to check if rewrites similar to the correct question will also produce correct answers, and vice versa.

\begin{figure}[h]
  \centering
  \includegraphics[width=\linewidth]{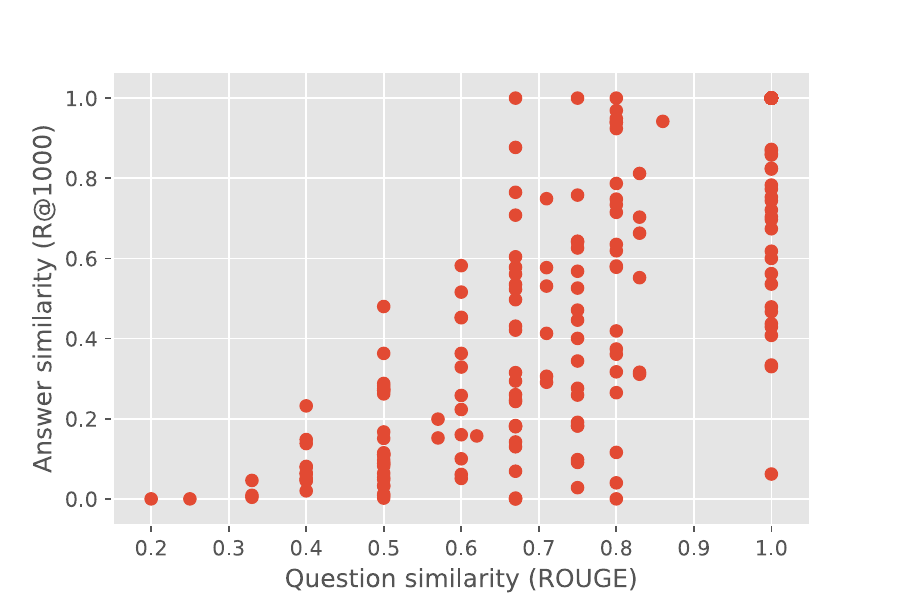}
  \caption{Strong correlation (Pearson 0.77) between question similarity (ROUGE) and the answer sets produced by the passage retrieval model (Recall).}
    \label{fig:rouge_recall}
\end{figure}

\begin{figure}[h]
  \centering
  \includegraphics[width=\linewidth]{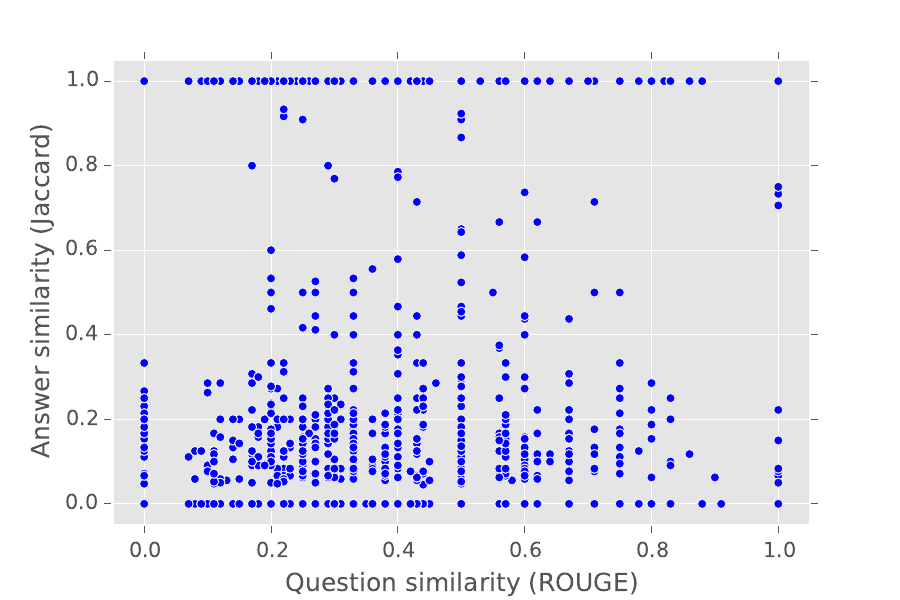}
  \caption{Weak correlation (Pearson 0.31) between question similarity (ROUGE) and answers produced by the reading comprehension model (Jaccard).}
    \label{fig:rouge_jaccard}
\end{figure}

There is a strong correlation for ROUGE = 1, i.e., when the generated rewrite is very close to the human one. However, when ROUGE $<$ 1 the answer is less predictable. Even for rewrites that have a relatively small lexical overlap with the ground-truth (ROUGE $\leq$ 0.4) it is possible to retrieve a correct answer, and vice versa.

We further explore the effect of the QR quality on the QA results by comparing differences of the answer sets produced when given different rewrites. We compare answers produced separately for human and model rewrites using the same input question. However, this time we look at all the answers produced by the QA model irrespective of whether the answers were correct or not. This setup allows us to better observe how much the change in the question formulation triggers the change in the produced answer.

Figure~\ref{fig:rouge_recall} demonstrates strong correlation between the question similarity, as measured by ROUGE, and the answer set similarity. We measured the similarity between the top-1000 answers returned for the human rewrites and the generated rewrites by computing recall (R@1000). Points in the bottom right of this plot show sensitivity of the QA component, where similar questions lead to different answer rankings. The data points that are close to the top center area indicate weaknesses of the QR metric as a proxy for the QA results: often questions do not have to be the same as the ground truth questions to solicit the same answers. The blank area in the top-left of the diagonal shows that a lexical overlap is required to produce the same answer set, which is likely due to the candidate filtering phase based on the bag-of-word representation matching.

We also compared ROUGE and Jaccard similarity for the tokens in reading comprehension results but they showed only a weak correlation (Pearson 0.31). This result confirms our observation that the extractive model tends to be rather sensitive to a slight input perturbation but will also often provide the same answer to very distinct questions.

Thus, our results show that the existing QR metrics have a limited capacity in predicting the performance on the end-to-end QA task. ROUGE correlates with the answer recall for the retrieval model, but cannot be used to reliably predict answer correctness for the individual question rewrites. Since ROUGE treats all the tokens equally (except for the stop-word list) it is unable to capture importance of the individual tokens that modify question semantics. The correlation of the QA performance with the embedding-based USE metric is even lower than with ROUGE for the both QA models.

\section{Discussion}
\label{sec:discussion}

We showed that QA results can identify question paraphrases. However, this property directly depends on the ability of the QA model to match equivalent questions to the same answer set.

\paragraph{Wrong answer or wrong question?}

Our passage retrieval model is using BM25 as a filtering step, which relies on the lexical match between the terms in the question and the terms in the passage. Hence, synonyms, like ``large" and ``big", cannot be matched with this model. This effect explains that dissimilar questions are never matched to the same answers in Figure~\ref{fig:rouge_recall}. The drawback of this model is that it suffers from the ``vocabulary mismatch’’ problem~\cite{van2017remedies}.

A one-word difference between a pair of questions may have a very little as well as a very dramatic effect on the question interpretation. This class of errors corresponds to the variance evident from Figure~\ref{fig:rouge_recall}.

Our error analysis indicates that small perturbations of the input question, such as anaphora resolution, often cause a considerable change in the answer ranking. For example, the pair of the original question: ``\textit{Who are the Hamilton Electors and what were \textbf{they} trying to do?}'', and the human rewrite: ``\textit{Who are the Hamilton Electors and what were the \textbf{Hamilton Electors} trying to do?}'' produce ROUGE = 1 but R@1000 = 0.33. We also identified many cases in which inability of the QR component to generate apostrophes resulted in incorrect answers (original question: ``\textit{Describe \textbf{Netflixs} subscriber growth over time}'', and the human rewrite: ``\textit{Describe \textbf{Netflix's} subscriber growth over time}'').

In contrast, our reading comprehension models is based solely on the dense vector representations, which should be able to deal with paraphrases. In practice, we see from Figure~\ref{fig:rouge_jaccard} that this feature may also introduce an important drawback, when the model produces a correct answer even when given an incorrectly formulated question. This stability and undersensitivity of the reading comprehension model may indicate biases in the dataset and evaluation setup. When the answers no longer depends on the question being asked, but can be also predicted independently from the question, the QA model evidently fails to learn the mechanisms underlying QA. Instead, it may learn a shortcut in the benchmark dataset that allows to guess correct questions, such as likely answer positioning within an article.

Our experiments show that the retrieval-based setup is more adequate in judging model robustness. The size of the answer space is sufficiently large so as to exclude spurious cues that the model can exploit for shortcut learning. The role of QR is, therefore, much more evident in this task, since to be able to find the correct answer the question has to be formulated well by disambiguating conversational context. While it may be redundant to overspecify the question when given several passages to choose the answer from, it becomes of a vital importance when given several million passages. This argument holds, however, only when assuming a uniform answer distribution, which is often not the case unless for sufficiently large web collections. This particular observation brings us to the next question that we would like to discuss in more detail.

\paragraph{Rewrite or not to rewrite?}

Our experiments demonstrated the challenges in the quality control of the QR task itself. Human rewrites are not perfect themselves since it is not always clear whether and what should be rewritten. Considering the human judgment of the QR quality independent from the QA results, little deviations may not seem important. However, from the pragmatic point of view, they may have a major impact on the overall performance of the end-to-end system.

The level of detail required to answer a particular question is often not apparent and depends on the dataset.
However, we can argue that inconsistencies, such as typos and paraphrases, should be handled by the QA component, since they do not originate from the context interpretation errors.

Further on, we evaluated our assumption about human rewrites as the reliable ground truth. 
Our evaluation results indeed showed that human rewriting was redundant in certain cases. There are cases in which original questions without rewriting were already sufficient to retrieve the correct answers from the passage collection (see last row of the Table~\ref{table:trec_breakdown}). In particular, we found that 10\% of the questions in TREC CAsT were rewritten by human annotators that did not need rewriting to retrieve the correct answer.  For example, original question: ``\textit{What is the functionalist theory?}'', human rewrite: ``\textit{What is the functionalist theory in \textbf{sociology}?}''. However, in another question from the same dialogue, omitting the same word from the rewrite leads to retrieval of an irrelevant passage, since there are multiple alternative answers.

The need for QR essentially interacts with the size and diversity of the possible answer space, e.g., collection content, with respect to the question. Some of the questions were correctly answered even with underspecified questions, e.g., original question: ``\textit{What are some ways to avoid injury?}'', human rewrite: ``\textit{What are some ways to avoid \textbf{sports} injuries?}'', because of the collection bias.

The goal of QR is to learn patterns that correct question formulation independent from the collection content. This approach is similar to how humans handle this task when formulating search queries, i.e., based on their knowledge of language and the world. Humans can spot ambiguity regardless of the information source by modeling the expected content of the information source even without the need to have direct access to it. Clearly, this expectation may not be optimal or even sufficient to be able to formulate a single precise question, which is exactly the point at which the need for an information-seeking dialogue naturally arises.

When the question formulation procedure is designed to be independent of the collection content, it also allows for querying several sources with the same question. This property is especially helpful when the content of the collection is unknown, which also allows to work with 3rd party APIs to access information in a distributed fashion. We may see a similar effect as in human-human communication here as well, when the need for an automated information-seeking dialogue between distributed systems may arise to better negotiate the information need and disambiguate the question further with respect to the content of each remote information source.

\section{Related Work}
\label{sec:related}




Several recent research studies reported that state-of-the-art machine learning models lack in robustness across several NLP tasks~\cite{DBLP:conf/aclnut/LiS19,DBLP:conf/cvpr/ZengLWQXTTY19}. In particular, these models were found to be sensitive to input perturbations.

\citet{DBLP:conf/emnlp/JiaL17} analysed this problem in the context of the reading comprehension task. They showed in the experimental evaluation that QA models suffer from overstability, i.e., they tend to provide the same answer to a different question, when it is sufficiently similar to the correct question. \citet{DBLP:conf/iclr/LewisF19} also showed this deficiency of the state-of-the-art models for reading comprehension that learn to attend to just a few words in the question.

Our evaluation results support these findings using the conversational settings. We showed that many ambiguous questions in QuAC can be answered correctly without the conversation history (see Table~\ref{table:quac_breakdown}). In contrast, this effect is absent in the passage retrieval setup (see Table~\ref{table:trec_breakdown} or Figure~\ref{fig:trend_plots}). These results suggest that the passage retrieval evaluation setup is more adequate for training robust QA models.

Previous work that focused on analysing and explaining performance of Transformer-based models on the ranking task used word attention visualisation and random removals of non-stop words~\cite{dai2019deeper,qiao2019understanding}.
The analysis framework proposed here provides a more systematic approach to model evaluation using ambiguous and rewritten question pairs, which is a by-product of applying QR in conversational QA.

Our approach is most similar to the one proposed by \citet{DBLP:conf/acl/SinghGR18}, who used semantically equivalent adversaries to analyse performance in several NLP tasks, including reading comprehension, visual QA and sentiment analysis. Similarly, in our approach, question rewrites, as paraphrases with different levels of ambiguity, are a natural choice for evaluation of the conversational QA performance. The difference is that we do not need to generate semantically equivalent questions but can reuse question rewrites as a by-product of introducing the QR model as a component of the conversational QA architecture.

\section{Conclusion}
\label{sec:conclusions}

QR is a challenging but a very insightful task designed to capture linguistic patterns that identify and resolve ambiguity in question formulation. Our results demonstrate the utility of QR as not only enabler for conversational QA but also as a tool that helps to understand when QA models fail.

We introduced an effective error analysis framework for conversational QA using QR and used it to evaluate sensitivity of two state-of-the-art QA model architectures (for reading comprehension and passage retrieval tasks). Moreover, the framework we introduced is agnostic to the model architecture and can be reused for performance evaluation of different models using other evaluation metrics as well.

QR helps to analyse model performance and discover their weaknesses, such as oversensitivity and undersensitivity to differences in question formulation. In particular, the reading comprehension task setup is inadequate to reflect the real performance of a question interpretation component since ambiguous or even incorrect question formulations are likely to result in a correct answer span. In passage retrieval, we observe an opposite effect. Since the space of possible answers is very large it is impossible to hit the correct answer by chance. We discover, however, that these models tend to suffer from oversensitivity instead, i.e., when even a single character will trigger a considerable change in answer ranking.

In future work, we should extend our evaluation to dense passage retrieval models and examine their performance using the conversational QA setup with the QR model. We should also look into training approaches that could allow QA models to further benefit from the QR component. As we showed in our experiments QR provides useful intermediate outputs that can be interpreted by humans and used for evaluation. We believe that QR models can be also useful for training more robust QA models. Both components can be trained jointly, which is inline with \citet{DBLP:conf/iclr/LewisF19}, who showed that the joint objective of question and answer generation further improves QA performance.


\section*{Acknowledgements}
We would like to thank our colleagues Srinivas Chappidi, Bjorn Hoffmeister, Stephan Peitz, Russ Webb, Drew Frank, and Chris DuBois for their insightful comments.

\bibliographystyle{ACM-Reference-Format}
\bibliography{main}

\end{document}